# Diffusion Models in Bioinformatics: A New Wave of Deep Learning Revolution in Action

Zhiye Guo[&], Jian Liu[&], Yanli Wang[&], Mengrui Chen[&], Duolin Wang, Dong Xu, Jianlin Cheng*

Department of Electrical Engineering and Computer Science, University of Missouri, Columbia, 65211, Missouri, USA

NextGen Precision Health, University of Missouri, Columbia, 65211, Missouri, USA

[&]Joint first authors; *corresponding author (chengji@missouri.edu)

**Abstract**

Denoising diffusion models have emerged as one of the most powerful generative models in recent years. They have achieved remarkable success in many fields, such as computer vision, natural language processing (NLP), and bioinformatics. Although there are a few excellent reviews on diffusion models and their applications in computer vision and NLP, there is a lack of an overview of their applications in bioinformatics. This review aims to provide a rather thorough overview of the applications of diffusion models in bioinformatics to aid their further development in bioinformatics and computational biology. We start with an introduction of the key concepts and theoretical foundations of three cornerstone diffusion modeling frameworks (denoising diffusion probabilistic models, noise-conditioned scoring networks, and stochastic differential equations), followed by a comprehensive description of diffusion models employed in the different domains of bioinformatics, including cryo-EM data enhancement, single-cell data analysis, protein design and generation, drug and small molecule design, and protein-ligand interaction. The review is concluded with a summary of the potential new development and applications of diffusion models in bioinformatics.

**Introduction**

Diffusion models are a kind of deep learning-based generative models [1-4] that aim to generate artificial yet realistic data (e.g., a computer-generated Picasso's painting) from input parameters. Various generative models such as Autoregressive Models (ARMs) [5], Normalizing Flows (NFs) [6], Energy-based Models (EBMs) [7], Variational Auto-Encoders (VAEs) [8], and Generative Adversarial Networks (GANs) [9] have been developed in the machine learning field. Particularly, GANs consisting of a generator of generating data and a discriminator of differentiating generated data from real data held a dominant position in generative tasks until recently.

Recently, diffusion-based generative models have become very popular in various domains and have many advantages over other generative models, including their ability to learn complex distributions smoothly, to handle high-dimensional data, and to generate extremely diverse data [10-15]. Diffusion models have surpassed the previously dominant generative adversarial networks (GANs) [9] in the challenging task of image synthesis [1,10]. They have achieved success in a variety of domains, ranging from computer vision [11,16-32], natural language processing (NLP) [15,33-36], temporal data modeling [37-42], multi-modal modeling [43-46], and life science-related domains, such as bioinformatics and medical image reconstruction [47-56]. Diffusion models were first introduced by Sohl-Dickstein et al. in [2]. Inspired by the concept of diffusion models, Ho et al. proposed Denoising Diffusion Probabilistic Models (DDPM) [1], for the first time showing that diffusion models can achieve performance comparable to other state-of-art generative models in image generation tasks. Following the work, Dhariwal and Nichol [10] further improved the diffusion network structure and the training strategy to boost the performance. The improved diffusion models surpassed GANs in image synthesis. Since then, diffusion models have emerged as the first choice for many generative tasks in different domains.

Most recently, numerous diffusion models started to be applied to address various bioinformatics problems, such as denoising cryo-EM data, single-cell gene expression analysis, protein design, drug and small molecule design, and protein-ligand interaction modeling. These models often outperform their predecessors, such as VAE and GAN, demonstrating a great potential of diffusion-based models in bioinformatics. This review provides a detailed survey of diffusion models and their current applications in bioinformatics. The main contributions of this review include:

- An accessible introduction of the essential techniques and tools of three generic diffusion models: denoising diffusion models, noise-conditioned score networks, and stochastic differential equations.
- A rather thorough survey of the applications of diffusion models in bioinformatics.
- A discussion of the potential future development of diffusion models in bioinformatics.

**The concept and foundation of diffusion models**

Diffusion models learn to reverse the process of data destruction/corruption (e.g., introducing noises to data), allowing for the generation of realistic, clean data samples (e.g., restoration of uncorrupted data). It offers a powerful approach to learning from data progressively destroyed or degraded. These models can be used to generate new samples from a given distribution or to estimate the distribution from which a given sample is drawn.

Despite diverse applications, the diffusion models to tackle them are mainly based on the three predominant formulations shown in **Figure 1**: (1) denoising diffusion probabilistic models (DDPMs) [1,2,57], (2) noise-conditioned score networks (NCSNs) [3,58], and (3) stochastic differential equations (SDEs) [4,59], which are described in details as follows.

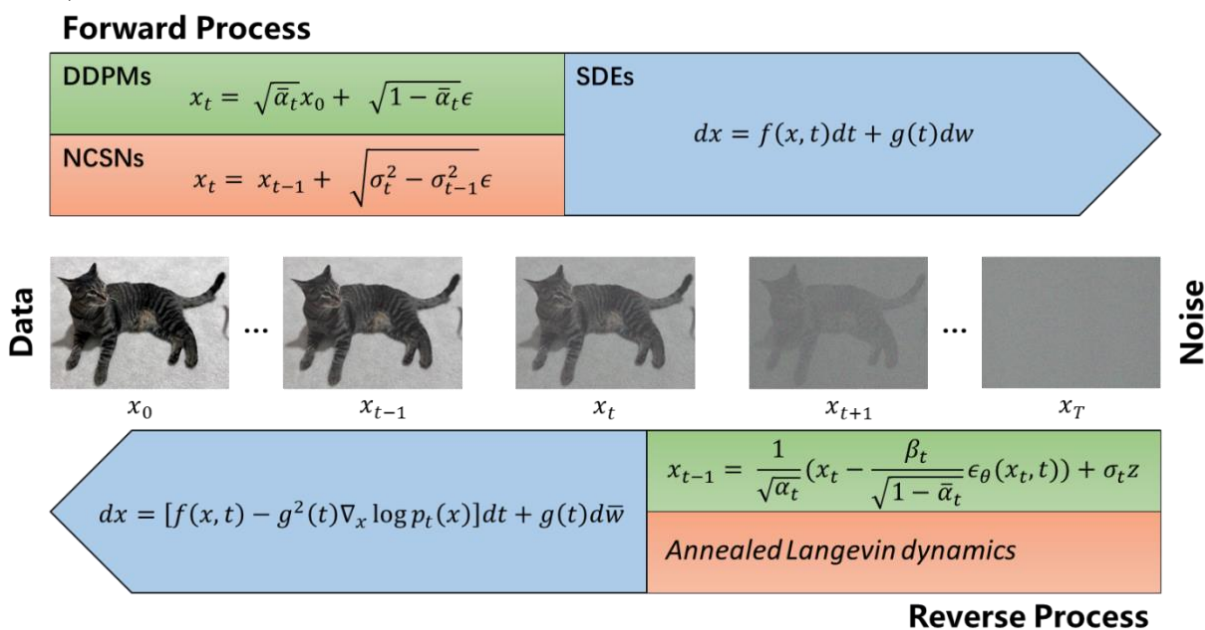

**Fig. 1.** A simplified illustration of the forward and reverse processes of three key diffusion models: denoising diffusion probabilistic models (DDPMs) in green, noise-conditioned score networks (NCSNs) in orange, and stochastic differential equations (SDEs) in blue. The forward process is to add noise into data (e.g., an image of a cat) progressively (from time 0 to T). The reverse process is to denoise the data from time T to 0.

**Denoising diffusion probabilistic models (DDPMs)**

The popularity of diffusion models mainly started from the work of DDPM [1], though the main mathematical framework of DDPM came from Sohl Dickstein's work in 2015 [2]. DDPM is the first diffusion model of generating high-resolution data. There are two Markov chains in a typical DDPM: the forward chain gradually adds noise to scramble the original data, followed by a reverse chain that removes the noise from the data to recover it to the original one. Let $q(x_0)$ be the distribution of the original data in which $x_0$ denotes the uncorrupted data. The transition kernel $q(x_t \mid x_{t-1})$ of the forward Markov process adding Gaussian perturbation at the time $t$ is denoted as $\mathcal{N}(x_t; \sqrt{1-\beta_t}\, x_{t-1}, \beta_t \mathbf{I})$, in which $t \in \{1, \ldots, T\}$. T represents the number of diffusion steps; $\beta_t \in [0,1)$ is the hyperparameter denoting the variance schedule across diffusion steps; $\mathbf{I}$ is the identity matrix; and $\mathcal{N}(x; \mu, \sigma)$ is the normal distribution of $x$ with mean $\mu$ and covariance $\sigma$. Let $\alpha_t = 1 - \beta_t$ and $\bar{\alpha}_t = \prod_{s=0}^{t} \alpha_s$, a noisy sample $(x_t)$ can be directly obtained from the distribution conditioned on the original input $x_0$ below:

$$q(x_t \mid x_0) = \mathcal{N}(x_t; \sqrt{\bar{\alpha}_t} x_0, (1-\bar{\alpha}_t)\mathbf{I}) \tag{1}$$

$$x_t = \sqrt{\bar{\alpha}_t} x_0 + \sqrt{1-\bar{\alpha}_t}\, \epsilon,\ \epsilon \sim \mathcal{N}(0, \mathbf{I}) \tag{2}$$

The forward process gradually introduces noise into the original data until it is completely replaced by noise. The reverse process is an opposite operation which is also the process of generating new samples that usually start with an unstructured noise obeying the prior distribution. It applies a model, usually a trainable neural network with the learning ability to remove noise step by step to restore the original data. The neural network *N* is formulated as follows:

$$p_\theta(x_{t-1} \mid x_t) = \mathcal{N}(x_{t-1}; \mu_\theta(x_t, t), \sigma_\theta(x_t, t)) \tag{3}$$

And given a starting point data of the reverse process as $p(X_T) = \mathcal{N}(X_T; 0, I)$, we get the distribution of $X_0$ conditioned on $X_T$:

$$p_\theta(X_{0:T}) = p(X_T) \prod_{t=1}^{T} p_\theta(X_{t-1} \mid X_t) \tag{4}$$

Eventually, a marginal distribution of $X_0$ close to the original data $x_0$ can be obtained by $p_\theta(x_0) = \int p_\theta(x_{0:T})\, dx_{1:T}$.

To train the model parameterized with $\theta$ to learn the pattern of the original data and make $p(x_0)$ close to the true data distribution $q(x_0)$, the loss function to be minimized is set as the negative log-likelihood below:

$$\begin{aligned}\mathbb{E}[-\log p_\theta(x_0)] &\leq \mathbb{E}_q\left[-\log \frac{p_\theta(x_{0:T})}{q(x_{1:T}\mid x_T)}\right] \\ &= \mathbb{E}_q\left[-\log p(x_T) - \sum_{t\geq 1} \log \frac{p_\theta(x_{t-1}\mid x_t)}{q(x_t\mid x_{t-1})}\right] \\ &= -L_{VLB}\end{aligned} \tag{5}$$

The objective of DDPM training is to minimize $L_{VLB}$, also known as the variational lower bound of the log-likelihood. Ho et al. [1] parametrized the $L_{VLB}$ to increase the quality of sample generation. The parameterized training objective is equivalent to that of the noise-conditional score networks (NCSNs) [3], which is described in the following sub-section.

**Noise-Conditioned Score Networks (NCSNs)**

In NCSNs, the score function of a probability density function $p(x)$ is represented by the gradient of the log density with respect to the input as $\nabla_x \log p(x)$. To learn and estimate the score function, a score-matching neural network $s_\theta$ is trained. The goal of this neural network is to make $s_\theta(x) \approx \nabla_x \log p(x)$. Therefore, the objective function of the scoring network can be defined as:

$$\mathbb{E}_{x \sim p(x)} \| s_\theta(x) - \nabla_x \log p(x) \|_2^2 \tag{7}$$

Even though the problem is well defined, optimizing Equation 7 is almost impossible numerically because there is no way of knowing the value of $\nabla_x \log p(x)$. However, there are some established techniques for learning score functions from data, such as score matching [60], denoising score matching [61-63], and sliced score matching [64].

Moreover, in work [3], the authors highlighted that the main difficulty of training lied in the fact that the trained score functions were unreliable in a low-dimension manifold because data usually lied in a low-dimension manifold embedded in a high-dimension space (the manifold hypothesis). They also showed that these issues could be solved by introducing Gaussian noise to the data at various scales, which improved the data distribution's suitability for the score-based generative modeling.

Hence, they proposed a single noise-conditioned score network (NCSN) to estimate the score corresponding to each noise level. Let $0 < \sigma_1 < \sigma_2 < \cdots < \sigma_t < \cdots < \sigma_T$ be a sequence of Gaussian noise levels, so that $p_{\sigma_t}(x_t|x) = \mathcal{N}(x_t; x, \sigma_t^2 \mathbf{I})$, $p_{\sigma_1}(x) \approx p(x_0)$, and $p_{\sigma_T}(x) \approx \mathcal{N}(0, \mathbf{I})$. The noise-conditioned score network $s_\theta(x, \sigma_t)$ with the denoising score matching can approximate the gradient log density function, making $s_\theta(x, \sigma_t) \approx \nabla_x \log\left(p_{\sigma_t}(x)\right), \forall\, t \in \{1, \ldots, T\}$. And for $x_t$, $\nabla_x \log(p_{\sigma_t}(x))$ is derived as:

$$\nabla_{x_t} \log p_{\sigma_t}(x_t|x) = -\frac{x_t - x}{\sigma_t} \tag{8}$$

Consequently, the optimization objective function in Eq. (7) can be transformed into:

$$\frac{1}{T} \sum_{t=1}^{T} \lambda(\sigma_t) \, \mathbb{E}_{p(x)} \mathbb{E}_{x_t \sim p_t(x_t|x)} \left\| s_\theta(x_t, \sigma_t) + \frac{x_t - x}{\sigma_t} \right\|_2^2 \tag{9}$$

in which the $\lambda(\sigma_t)$ is a weighting function.

During the sampling phase, NCSNs uses the annealed Langevin dynamics algorithm, which employs a Markov Chain Monte Carlo (MCMC) procedure to simply sample from a distribution according to its score function $\nabla_x \log p(x)$. The Langevin method recursively computes $x_i$ as follows:

$$x_i = x_{i-1} + \frac{\gamma}{2} \nabla_x \log p(x) + \sqrt{\gamma}\, \omega_i \tag{10}$$

where $\gamma$ determines the amplitude of the update in the score's direction; $x_0$ is sampled from the prior distribution; and the noise is drawn according to $\omega_i \sim \mathcal{N}(0, \mathbf{I})$.

**Stochastic differential equations (SDEs)**

With unlimited time steps or noise levels, DDPMs and NCSNs can be further generalized to the situation in which the perturbation and denoising processes can be described as stochastic differential equations (SDEs). This generalized approach [4] of gradually transforming the data into noise is called formulation score SDE. The forward process of the formulation score SDE uses stochastic differential equations and requires an estimated score function of the noisy data distribution. It is equivalent to the Itô SDE [65] solution, which consists of a drift component for mean transformation and a diffusion coefficient for describing noise:

$$dx = f(x, t)dt + g(t)dw, \quad t \in [0, T] \tag{11}$$

where $w$ represents the standard Wiener process known as Brownian motion, and $f(x, t)$ and $g(t)$ are the drift and diffusion coefficients of SDE. The forward process in DDPMs and SGMs is a special case of the discretizational SDE.

The formulation of the reverse diffusion process of SDE is given by Equation 12 [66], also called reverse-time SDE:

$$dx = [f(x, t) - g^2(t) \nabla_x \log p_t(x)]dt + g(t)d\overline{w} \tag{12}$$

where $\overline{w}$ is the standard Brownian motion running backward time, and $dt$ represents the infinitesimal negative time step. The reverse SDE and forward SDE share the same marginal densities but in the opposite time direction [4]. Like DDPMs and NCSNs, to numerically solve the reverse-time SDE, a trainable neural network $s_\theta(x, t)$ is employed to estimate the actual score function $\nabla_x \log p_t(x)$. The objective function can be defined as:

$$\mathbb{E}_{x(t) \sim p(x(t)|x(0)), x(0) \sim p_{data}} \left[ \frac{\lambda(t)}{2} \left\| s_\theta(x(t), t) - \nabla_{x(t)} \log p_t(x(t)|x(0)) \right\|_2^2 \right] \tag{13}$$

where $t \sim \mathcal{U}([0, T])$ denotes the uniform distribution over $[0, T]$ and $\lambda$ is a weighting function. In addition, Song et al. proposed several sampling techniques, like the Predictor-Corrector sampler, to generate good samples. This procedure uses a score-based method (i.e., annealed Langevin dynamics) as a corrector after using a numerical approach to sample data from the reverse-time SDE.

**Implementation of diffusion models**

In recent years, there have been some practical implementations of the foundational diffusion models and many improvements or extension in training speed, data sampling, integration with other neural networks, and applications to different data types, which make diffusion models useful in practice. The open-source tools that implement or improve various diffusion models are listed in **Table 1**.

**Table 1**. Open-source tools of implementing the foundational diffusion models or improving the usability of the diffusion models. The number of GitHub stars was recorded on 01/12/2023.

| Tool Name | Year | Model type or improvement | Architecture | GitHub | # of stars |
|---|---|---|---|---|---|
| DDPM [1] | 2020 | Foundational model | DDPM | CODE | 1500 |
| NCSN [3] | 2019 | Foundational model | NCSN | CODE | 456 |
| Score SDE [4] | 2020 | Foundational model | Score SDE | CODE | 820 |
| Diffusion Distillation [67] | 2022 | speed up training | DDPM | CODE | 26700 |
| ARDM [68] | 2022 | categorical data | DDPM | CODE | 26700 |
| Improved Diffusion [57] | 2021 | speed up training | DDPM | CODE | 1300 |
| k-Diffusion [69] | 2022 | speed up sampling | Score SDE | CODE | 856 |
| Cold Diffusion [70] | 2021 | speed up training | Score SDE | CODE | 690 |
| DPM-Solver [71] | 2022 | speed up sampling | Score SDE | CODE | 678 |
| VQ-diffusion [72] | 2022 | vector quantized | DDPM | CODE | 602 |
| DDIM [73] | 2021 | speed up sampling | DDPM | CODE | 470 |
| Diffusion-point-cloud [74] | 2021 | point cloud | DDPM | CODE | 331 |
| Improved VQ-Diff [75] | 2022 | vector quantized | DDPM | CODE | 295 |
| Diffusion GAN [76] | 2022 | mixed modeling | DDPM | CODE | 261 |
| LSGM [77] | 2021 | mixed modeling | Score SDE | CODE | 244 |
| DiffuseVAE [78] | 2022 | mixed modeling | DDPM | CODE | 226 |
| PNDM [79] | 2022 | speed up sampling | Score SDE | CODE | 200 |
| ConfGF [80] | 2021 | diffusion on graph | Score SDE | CODE | 177 |
| GeoDiff [81] | 2022 | diffusion on graph | DDPM | CODE | 161 |
| VDM [82] | 2022 | speed up training | DDPM | CODE | 146 |
| Analytic-DPM [83] | 2022 | speed up sampling | NCSN | CODE | 120 |
| PVD [84] | 2021 | point cloud | DDPM | CODE | 108 |
| sdeflow-light [85] | 2021 | SDE unification | Score SDE | CODE | 90 |
| Gotta Go Fast [86] | 2021 | speed up sampling | Score SDE | CODE | 89 |
| Score-flow [59] | 2021 | mixed modeling | Score SDE | CODE | 84 |
| DiffFlow [87] | 2021 | mixed modeling | Score SDE | CODE | 77 |
| FastDPM [88] | 2021 | speed up training | Score SDE | CODE | 69 |
| Point Diffusion-Refinement [13] | 2022 | point cloud | DDPM | CODE | 69 |
| EDP-GNN [89] | 2020 | diffusion on graph | NCSN | CODE | 64 |
| GDSS [90] | 2022 | diffusion on graph | Score SDE | CODE | 55 |
| argmax_flows [34] | 2021 | categorical data | DDPM | CODE | 54 |
| riemannian-score-sde [91] | 2022 | diffusion on manifold | Score SDE | CODE | 44 |
| Soft Truncation [92] | 2021 | likelihood optimization | Score SDE | CODE | 42 |

**Applications of diffusion models in bioinformatics**

We group the applications of diffusion models in bioinformatics into five categories according to the types of problems (**Figure 2**). For each category, we first introduce the problem and then describe how specialized diffusion models have been applied to it in the following sub-sections.

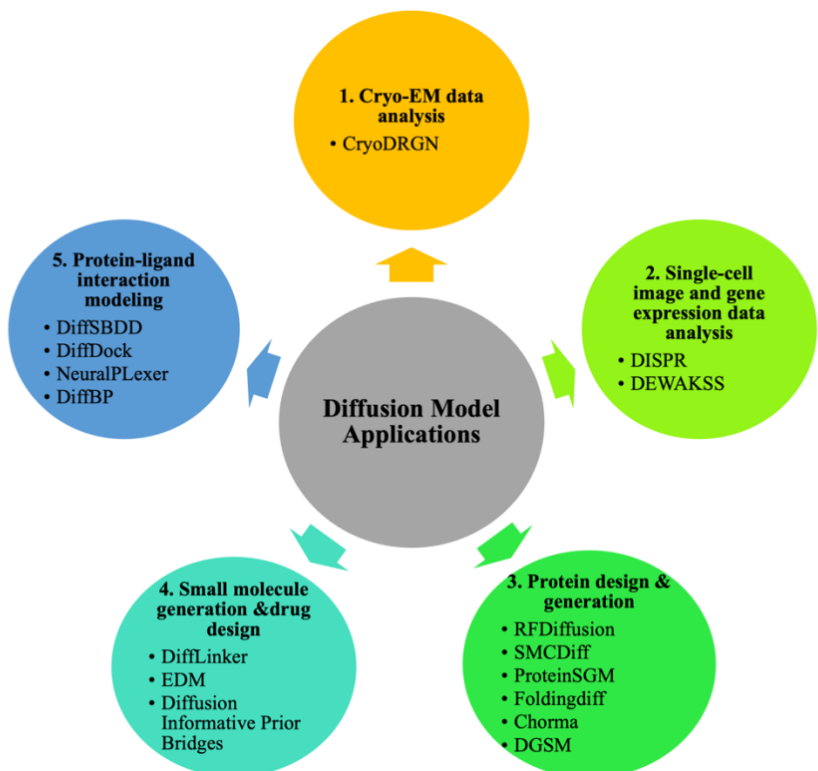

**Fig. 2**. The five types of applications of diffusion models in bioinformatics, including some representative tools and their references. (1) Cryo-EM data analysis, (2) single-cell image and gene expression data analysis, (3) protein design and generation, (4) small molecule generation and drug design, and (5) protein-ligand interaction modeling.

**Cryo-EM data analysis**

Single particle cryo-electron microscopy (cryo-EM) is a key imaging technique to determine and visualize the 3D conformation (structure) of large biomolecular complexes (e.g., protein complexes) at the atomic resolution. The images of protein complexes obtained by cryo-EM can be used to reconstruct their 3D conformation represented by 3D density maps.

A protein complex structure reconstruction method – CryoDRGN [93] introduced a latent variable $Z$ to define a conformational space $V$ for a protein complex on cryo-EM density maps. The traditional CryoDRGN based on Variational Autoencoder (VAE) framework learns a continuous distribution in the latent space for protein structures from cryo-EM data. Though CryoDRGN can simulate complicated structural dynamics, the Gaussian prior distribution of VAE does not match the aggregate approximation posterior, which limits the generative capability of model. A continuous-time diffusion model (i.e., Score SDEs) was recently introduced into CryoDRGN to learn a high-quality generative model for protein conformations directly from cryo-EM imaging data. The new CryoDRGN [94], equipped with the expressive denoising diffusion generative models, can better capture the 3D conformations of protein complexes and reconstruct 3D conformation of better quality.

**Single-cell image and gene expression data analysis**

Reconstructing the 3D shape of a single cell from an individual single-cell 2D microscopy image is useful for studying the morphological features of cells. Computational methods for addressing this problem face a significant challenge: each single 2D image may permit multiple 3D reconstructions, while randomly selecting a 2D slice can lead to a different prediction of the 3D shape. To tackle the uncertainty issue, DISPR [95] uses a 2D image of an individual cell as an inductive bias, and then constrains the diffusion-based model with the 2D image to predict 3D cell shapes, which builds realistic 3D shapes of the cell from the 2D image. Training a stochastic model to predict the infinite number of probable reconstructions rather than obtaining a single, deterministic reconstruction makes DISPR highly successful. It is the first use of a diffusion model in the context of 3D shape reconstruction for cells.

Single-cell RNA sequencing (scRNA-seq) can measure the expression of the genes in individual cells. However, the extremely low quantities of RNAs presented in an individual cell lead to highly noisy measurements of gene expression, especially missing values (dropouts). Therefore, it is important to denoise scRNA-seq data and impute the missing values. DEWAKSS [96] applies a diffusion model with K-Nearest Neighbor Graph (KNN-G) to select denoising hyperparameters via the noise2self self-supervision method, not depending on an explicit noise model but on an invariant function of the data features. Different than heuristic-based methods causing over-smoothed data variance, DEWAKSS can maintain variances along several gene expression dimensions, which makes it more competitive than the previous methods.

**Protein design and generation**

Given a protein structure, designing protein sequences that can fold into the structure (i.e., inverse protein folding problem) is important for designing novel proteins, such as enzymes, to carry out some specific functions more effectively than natural proteins. RFDiffusion [97] constructs a DDPM generative model of protein backbones, which can be used to design protein monomers, symmetric protein oligomers, enzyme active site scaffolding, and symmetric motif scaffolding. It has been used to design therapeutic and metal-binding proteins. Some proteins designed by RFDiffusion have been experimentally validated.

Another diffusion model SMCDiff [98] was also proposed to address the motif-scaffolding problem of designing a scaffold structure given a protein motif. It first uses the diffusion-based model ProtDiff to predict the protein backbone and then applies the SMCDiff to generate scaffolds for arbitrary motifs. It can generate longer and more diverse scaffolds than the previous methods.

ProteinSGM [54] is a score-based generative diffusion model, which generates images-like 2D matrices consisting of inter-residue pairwise distances and angles to represent protein backbone structures. The matrices are then used to generate native-like protein3D structures, which can be used to design novel proteins. Similarly, Foldingdiff [99] applies a DDPM model with a vanilla transformer model to generate protein sequences that fold into a backbone protein structure. Some generated protein sequences were tested by protein structure prediction tools (e.g., AlphaFold2 [100]) to check if they could fold into the targeted protein structure, even though the folds of the designed protein sequences had not been verified by experiments, such as x-ray crystallography.

Chorma [101] is a diffusion-based generative model for generating large single proteins (>3000 residues) and protein complexes. The model can take different constraints, such as residue-residue distances, symmetry, and shape to control the structure generation. It uses a backbone diffusion network to capture the process of turning a collapsed polymer system into a protein complex backbone, and a graph-based design network to generate protein sequences conditionally on the sampled backbone. Compared to some previous work, it adopts a random graph neural network to reduce the time complexity from $O(N^2)$ or $O(N^3)$ down to $O(N)$, enabling it to generate large proteins and complexes efficiently.

A generative model designed in work [47] introduces a fully data-driven DDPM [1] model for generating realistic proteins across the full range of structural domains in the Protein Data Bank (PDB) [102]. Its invariant point attention modules achieve equivariance to rotations and translations of protein structures. It employs an approach similar to the masked language modeling to generate discrete protein sequences. Different from some previous methods, it can produce large proteins with multiple domain topologies.

Dynamic Graph Score Matching (DGSM) [103] was designed to predict stable 3D conformations from 2D molecular graphs mainly in computational chemistry and then extended to protein sidechain conformation prediction and multi-molecular complex prediction. It can model both local and long-range interactions by dynamically building graph structures according to the atom-atom spatial proximity. Especially, using the score matching method, the model can directly estimate the gradient fields of the logarithm density of atomic coordinates. The model can be trained in an end-to-end fashion. The new model can largely fix the weakness of overlooking the long-range interactions between non-bounded atoms in traditional experimental and physics-based simulation methods.

**Small molecule generation and drug design**

Designing small molecules (e.g., drugs) to mediate protein structure and function is important for both basic biomedical research and drug design. Fragment-based drug design is one of the strategies for discovering new small molecules, such as drug candidates, in 3D space. Given molecular fragments, the goal is to design linkers consisting of atoms to connect the fragments into a complete, viable molecule. DiffLinker [104] uses an E(3)-equivariant 3D-conditional diffusion model to generate molecular linkers to connect the molecular fragments. It has two parts, a graph neural network to predict the size of the linker and an equivariant diffusion model to generate the linker between the input fragments. It can generate not only linkers for multiple fragments but also the number of atoms in the linker and the attachment points for the input fragments.

E(3)-Equivariant Diffusion Model (EDM) [14] performs the diffusion process on the atom coordinates and atom types in the Euclidean space to generate small molecules. It can generate molecule structures with up to 29 atoms, which is larger than the 9 heavy atoms limit of the previous methods. It combines the equivariant graph neural network (EGNN) and the diffusion process. The former models the molecule structures with geometric symmetries, while the latter makes training easier and enhances performance and scalability. Inspired by the physics governing the formation of small molecules, a simple but novel diffusion-based generative model based on the physical and statistics prior information (Diffusion Informative Prior Bridge) was designed in work [105] to guide the diffusion process to generate high-quality and realistic molecules. Several energy functions are integrated with the physical and statistical prior information to improve both molecule generation and uniformity-promoted 3D point cloud generation.

**Protein-ligand interaction modeling**

Predicting the pose (conformation) of a ligand bound with a protein is important for studying protein-ligand interaction, protein function, and discovering drugs. Different from traditional energy-minimization methods of docking a ligand against a protein, DiffSBDD [106] adopts an E(3)-equivariant 3D-conditional diffusion model to generate novel ligands conditioned on the protein pockets. Ligands are predicted using two strategies: (1) protein-conditioned generation and (2) ligand-inpainting generation. The former produces the new ligands given a fixed protein pocket; and the latter learns the joint distribution of the ligand and protein pocket first, and then the ligand is inpainted during the inference time. DiffDock [107] formulates a diffusion process over ligand poses. DiffDock has both a trained scoring model and a trained confidence model. Both of them are built on top of SE(3)-equivariant convolutional networks. The former is used to generate different poses of the ligand, and the latter is used to select the ligand poses with the highest confidence score, similar as in AlphaFold2 [100] for protein structure prediction.

NeuralPLexer [108] is a deep generative network that leverages the stochastic differential equation (SDE) to predict protein-ligand complex structures. The key component in the model is the equivariant structure diffusion module (ESDM), which predicts the atomic coordinates on a heterogeneous graph formed by protein atoms, ligand atoms, protein backbone frames, and ligand local frames. Using SDE, the model can handle unbound or predicted protein structure inputs and automatically accommodate the changes of protein structures when they bind with ligands.

DiffBP [109] was designed to generate ligands capable of binding to a specific protein. It adopts the diffusion denoising generative models in conjunction with equivariant graph neural networks [110,111] to generate high-quality ligand candidates [1,4,112]. Different from the previous methods of generating one atom at a time without considering the interactions among all atoms, DiffBP can generate ligands with the awareness of all the ligand atoms and the target protein.

**Bioinformatics diffusion model tools**

Some diffusion models applied to bioinformatics above have been implemented as open-source tools (**Table 2**), which are useful for users to apply them to their research problems and for developers to further improve them.

**Table 2.** Some open-source diffusion model tools for bioinformatics.

| Bioinformatics Domain | Tool Name | Denoising Condition | Architecture | GitHub Code |
|---|---|---|---|---|
| Cryo-EM data analysis | CryoDRGN | conditioned | Score SDE | CODE |
| Single-cell image and gene expression data analysis | DISPR | conditioned | Improved DDPM | CODE |
| Protein design and generation | FoldingDiff | conditioned and unconditioned | DDPM | CODE |
|  | Chorma | conditioned | Score SDE | CODE |
| Small molecule generation and drug design | DiffLinker | conditioned | Score SDE | CODE |
|  | EDM | conditioned | Score SDE | CODE |
| Protein-ligand interaction modeling | DiffSBDD | conditioned | DDPM | CODE |
|  | DiffDock | conditioned | DDPM | CODE |

**Future Development**

As described above, diffusion models have made an inroad into several important bioinformatics domains and significantly advance the state of the art. Nevertheless, we believe the potential of diffusion models in bioinformatics are far from being fully explored. Diffusion models will continue to rapidly expand their applications in bioinformatics due to their superior capabilities of denoising data and generating realistic new data in comparison with other generative models. The bioinformatics field has a rich set of such problems for diffusion models to tackle. Below are several examples to illustrate their potential future development in the field.

*3D genomics data analysis.* Hi-C data captures the interactions between chromosomal regions of a genome, which can be used to build 3D conformations of the genome [113,114] and study long-range gene-enhancer interactions. Hi-C sequence reads data are usually converted into 2D chromosomal contact maps. However, Hi-C data, particularly single-cell Hi-C data, are usually noisy and incomplete. Even though some deep learning methods (e.g., GANs) have been applied to denoising Hi-C data [115], diffusion models provide a new powerful new means to denoise Hi-C chromosomal contact maps to improve 3D genome conformation modeling and study spatial interactions between genes and regulatory elements (e.g., enhancers).

*Single-cell reconstruction and inference.* In addition to imputing and denoising gene expression values in single-cell RNA-seq data, various other single-cell problems may benefit from diffusion models: for example, inferring data of one modality (e.g., RNA-Seq) from another (e.g., ATAC-Seq), imputing missing spots in single-cell spatial transcriptomic data, decomposing spots (each consisting of multiple cells) in 10X spatial transcriptomic data into single cells (super-resolution), and building 3D models of cells' spatial arrangements using single-cell data.

*DNA regulatory element design.* Generative models such as generative adversarial networks (GAN) [116] have been applied to design enhancers to regulate the expression of genes and development of cell types. As diffusion models have shown better performance in image synthesis than GAN, it is expected that they may further advance the state of the art for designing enhancers.

*Cryo-EM image denoising.* Diffusion models have been applied to reconstruct protein complex structures from cryo-EM density maps as described in Section 3.1. As cryo-EM density maps are constructed from cryo-EM protein particle images that are also noisy, there is a significant need to denoise original cryo-EM images too. Even though traditional image preprocessing techniques have been applied to denoising cryo-EM images [117], diffusion models can provide some unique new capabilities to further advance the field.

*Peptide design.* As surveyed in this article, numerous diffusion models have been developed to design and generate new proteins. A related but different problem is to design peptides that can modulate protein function and even serve as drugs to treat disease. Therefore, similar diffusion models for protein design can probably be extended to design peptides.

*Protein structure prediction, refinement, mutation and function.* Stimulated by the successful applications of diffusion models to protein design and generation described in Section 3.3, it is expected more diffusion models will be developed to tackle other aspects of protein bioinformatics, such as refining predicted protein structures to make them closer to native structures, predicting how mutations change protein structures, and inferring protein functions.

*Proteomics and metabolomics data analysis.* A challenging problem in proteomics or metabolomics data analysis is to identify peptides or metabolites from mass spectrometry data. As mass spectrometry data is noisy, diffusion models are naturally fit for denoising such data. Moreover, diffusion models may be used to generate peptide sequences or metabolite structures from the mass spectrometry data to improve the identification of peptides and metabolites.

The potential applications above are just several example problems for diffusion models. It is expected that the bioinformatics and machine learning communities will find many creative ways to apply diffusion models to solve various bioinformatics problems not mentioned here in the next few years. Potentially, any bioinformatics problems that were handled using early generative methods, such as GAN and VAE, may be improved by diffusion models. However, it is worth mentioning that diffusion models also have some limitations. In particular, the computational resource requirements are much higher than GAN and VAE. Hence, it is important to evaluate the trade-off between performance improvement and computational resource demand before applying diffusion models. Furthermore, new applications of diffusion models are often non-trivial. The developers may need to extensively explore suitable data representations (embeddings), types of diffusion models, and deep learning architectures before achieving desirable results.

**Conclusion**

Diffusion models are igniting a new wave of deep learning application revolution in bioinformatics. To aid the community in quickly expanding the applications of diffusion models in the field, we provide a brief introduction of the concept and theoretical foundation of three main diffusion models

(DDPMs, NCSNs, and SDEs) and an overview of their applications in cryo-EM data analysis, single-cell data analysis, protein design, small molecule and drug design, and protein-ligand interaction modeling. Inspired by the exciting development, we envision many new bioinformatics applications in genomics, proteomics, metabolomics, and protein structure and function prediction for diffusion models to tackle. We expect many diffusion models will be developed in the next several years to significantly advance many areas of bioinformatics.

**Acknowledgements**

The work was partly supported by the National Institutes of Health (grant #.: R01GM146340 (JC), R01GM093123 (JC), and R35GM126985 (DX)), Department of Energy (grant #.: DE-SC0020400 (JC) and DE-SC0021303 (JC)), and National Science Foundation (grant #: DBI1759934 (JC) and IIS1763246 (JC)) of USA.

**Author Contributions Statement**

ZG, JL, YLW, and MRC collected the data. JC, DX, and DLW provided guidance on organizing the content. JC envisioned the new development in proteins/3D genomics/single-cell Hi-C data analytics/cryo-EM/DNA design/peptide/proteomics/metabolomics and XD envisioned the new development in single-cell reconstruction and inference. ZG, JC, JL, YLW, DX, DLW and MRC wrote/edited the manuscript.

**Competing Interests Statement**

The authors declare no competing interests.